\documentclass[10pt,twocolumn,letterpaper]{article}

\usepackage{cvpr}              

\usepackage{graphicx}
\usepackage{amsmath}
\usepackage{amssymb}
\usepackage{booktabs}
\usepackage{graphicx}  
\usepackage{subcaption}  
\usepackage{array}
\usepackage{enumitem}
\usepackage{multirow}
\usepackage{xcolor}
\usepackage{pifont}
\usepackage{makecell}
\usepackage{hyperref}

\usepackage{colortbl}
\usepackage{xcolor}
\definecolor{lightgreen}{rgb}{0.88, 1, 0.88}
\definecolor{lightred}{rgb}{1, 0.88, 0.88}
\usepackage{placeins}

\usepackage[capitalize]{cleveref}
\crefname{section}{Sec.}{Secs.}
\Crefname{section}{Section}{Sections}
\Crefname{table}{Table}{Tables}
\crefname{table}{Tab.}{Tabs.}

\def\cvprPaperID{33} 
\def\confName{CV4Animals}
\def\confYear{2025}

\begin{document}

\title{MMLA: \underline{M}ulti-Environment, \underline{M}ulti-Species, \underline{L}ow-\underline{A}ltitude Drone Dataset}

\author{
Jenna Kline\textsuperscript{1}, 
\and
  Samuel Stevens\textsuperscript{1},
  \and
  Guy Maalouf\textsuperscript{2}, 
  \and
  Camille Rondeau Saint-Jean\textsuperscript{2},
  \and
  Dat Nguyen Ngoc\textsuperscript{3}, 
  \and
  Majid Mirmehdi\textsuperscript{3}, 
  \and
  David Guerin\textsuperscript{4}, 
  \and
  Tilo Burghardt\textsuperscript{3}, 
  \and
  Elzbieta Pastucha\textsuperscript{2}, 
  \and
  Blair Costelloe\textsuperscript{5},
  \and
  Matthew Watson\textsuperscript{3}, 
  \and
  Thomas Richardson\textsuperscript{3}, 
  \and
  Ulrik Pagh Schultz Lundquist\textsuperscript{2} \and
  \footnotesize
  \textsuperscript{1}The Ohio State University
  \textsuperscript{2}University of Southern Denmark
  \textsuperscript{3} University of Bristol
  \textsuperscript{4} WildDroneEU
  \textsuperscript{5} Max Planck Institute of Animal Behavior
  }

\twocolumn[{%
  \maketitle
  \vspace{-1.5em}
  \begin{center}
    \includegraphics[width=\linewidth]{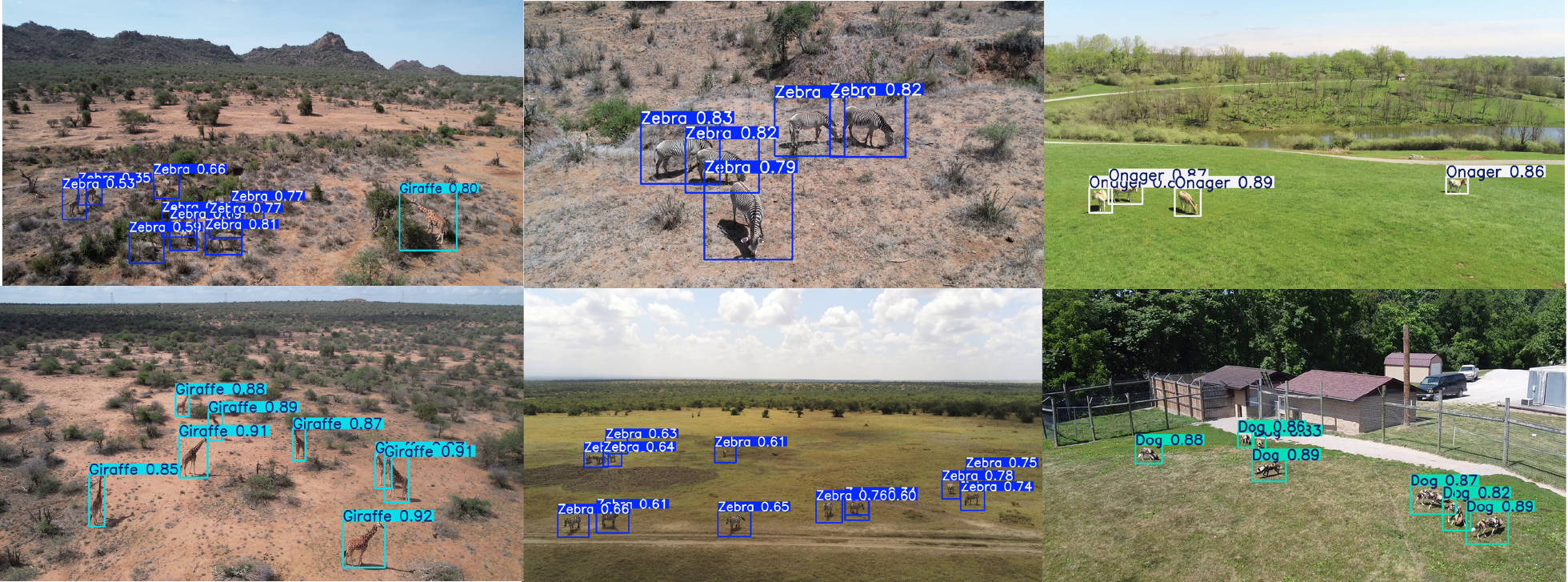}
    \captionof{figure}{Examples of the MMLA dataset and example detection results from a YOLOv11m model fine-tuned on MMLA. The dataset spans multiple species and environments, enabling robust model training for low-altitude wildlife detection.}
    \label{fig:enter-label}
  \end{center}
}]

\begin{abstract}
  Real-time wildlife detection in drone imagery supports critical ecological and conservation monitoring. However, standard detection models like YOLO often fail to generalize across locations and struggle with rare species, limiting their use in automated drone deployments. We present MMLA, a novel multi-environment, multi-species, low-altitude drone dataset collected across three sites (Ol Pejeta Conservancy and Mpala Research Centre in Kenya, and The Wilds in Ohio), featuring six species (zebras, giraffes, onagers, and African wild dogs). The dataset contains 811K annotations from 37 high-resolution videos. Baseline YOLO models show performance disparities across locations while fine-tuning YOLOv11m on MMLA improves mAP50 to 82\%, a 52-point gain over baseline. Our results underscore the need for diverse training data to enable robust animal detection in autonomous drone systems.
\end{abstract}


\section{Introduction}
\label{sec:intro}

Animal ecology studies increasingly rely on vision-based technologies for efficient, noninvasive wildlife monitoring. Camera traps, high-altitude and low-altitude drones each offer distinct advantages for different research goals. 
Machine learning models tailored for animal ecology data can rapidly analyze these datasets to produce valuable insights \cite{tuia2022perspectives}, but their effectiveness varies significantly across monitoring approaches.
While camera traps and high-altitude drone surveys have established methodologies and datasets \cite{orinoquia, wang_surveying_2019}, low-altitude drone studies represent an untapped opportunity. 
These deployments offer more detailed imagery than high-altitude approaches, covering broader areas than stationary camera traps and providing valuable insights into animal interactions and decision-making \cite{Koger_quantifying_2024}. 
Critically, low-altitude drones are increasingly deployed as autonomous systems that can actively track wildlife \cite{kline2025wildwing, kline2024integrating}, requiring models with high precision and localization.

While effective on benchmark datasets, standard computer vision models like YOLO \cite{jocher2023ultralytics} perform poorly when applied to wildlife detection in low-altitude drone imagery across varied natural environments. This performance gap stems from the specific requirements of autonomous wildlife-tracking drones: high precision detection with minimal false positives and accurate object localization (measured by IoU) are essential, as false detections or imprecise bounding boxes can cause autonomous systems to make incorrect flight decisions \cite{kline2024integrating}. Additionally, environmental variability across field sites introduces substantial domain shifts, causing models trained in one location to fail when deployed elsewhere.

To address this gap, we present a novel \textbf{\underline{m}ulti-environment, \underline{m}ulti-species, \underline{l}ow-\underline{a}ltitude (MMLA) drone dataset} \footnote{\href{https://huggingface.co/collections/imageomics/mmla-67f572d3ba17fca922c80182}{MMLA - an Imageomics HuggingFace Collection}}, the first low-altitude dataset explicitly designed for developing robust computer vision models for autonomous wildlife-tracking drones. Our dataset spans three distinct locations (Ol Pejeta Conservancy and Mpala Research Centre in Kenya and The Wilds Conservation Center in Ohio, USA) and six species: plains zebras, Grevy's zebras, reticulated and Masai giraffes, Persian onagers, and African wild dogs. 
MMLA contains 155K frames (1 hr, 26 min), with manually annotated bounding boxes around each animal optimized for training high-precision detection models.

The creation of MMLA represents a significant investment in advancing wildlife conservation technology, involving over 150 person-hours of fieldwork across two continents. By releasing this dataset and fine-tuned model weights, we invite the computer vision community to tackle these challenging real-world ecological problems and advance the state of the art in autonomous wildlife monitoring technology.

\section{Background \& Related Work}
\label{relatedwork}
\begin{table}[h!]
\centering
\small
\begin{tabular}{llll}
\toprule
\textbf{Feature} & \textbf{Camera} & \textbf{High-alt.} & \textbf{Low-alt.} \\
 & \textbf{Trap} & \textbf{Drone} & \textbf{Drone} \\
\midrule
Size\textsuperscript{1} & med-large & v. small & med \\
 & (128-256) & (0-16) & (64-128) \\
\midrule
Processing & Post-hoc & Post-hoc & Real-time \\
 & batches & batches & video \\
\midrule
Metric & Low false & F1-score & High prec. \\
 & negative &  & \& mAP50 \\
\midrule
Vision task & Detect, & Counting, & Localization \\
 & Species ID & Species ID &  \\
\midrule
Models & MegaDetector & POLO \cite{may2024polo}, & \textbf{None} \\
 & \cite{beery2019efficient} & HerdNet \cite{delplanque2023crowd} &  \\
\bottomrule
\end{tabular}
\caption{Comparison of vision-based animal ecology monitoring approaches. Low-altitude drone studies differ from camera trap and high-altitude drone studies for processing, relevant metrics, and computer vision tasks. \textbf{There are no pre-trained models aimed at low-altitude drone studies.} \\
\footnotesize\textsuperscript{1}Typical animal size in pixels \cite{he_longtailed_2023}.}
\label{tab:comparison}
\end{table}
\subsection{Monitoring Approaches for Animal Ecology}

\begin{figure}
    \centering
    \includegraphics[width=1\linewidth]{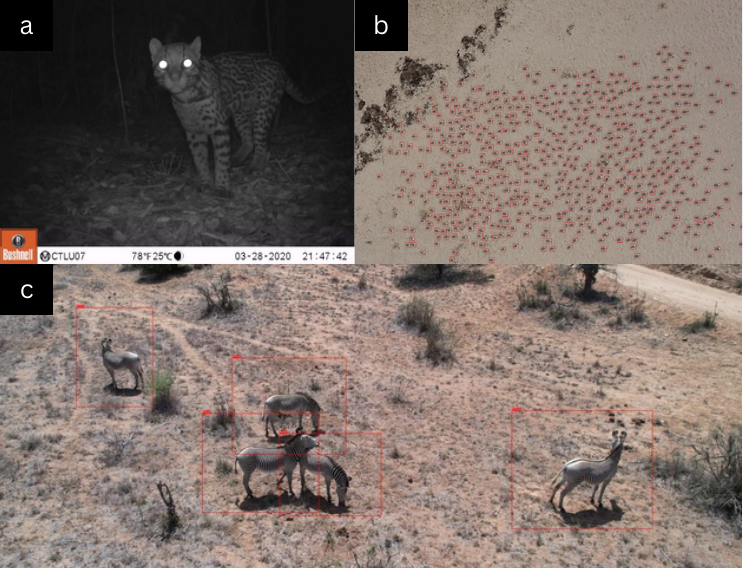}
    \caption{Animal imagery collected via (a) camera traps \cite{orinoquia}, (b) high-altitude drone \cite{dronescount}, and (c) low-altitude drone missions \cite{kholiavchenko2024kabr}.}
    \label{fig:topfigure}
\end{figure}

Vision-based wildlife monitoring approaches differ significantly in their technical specifications, applications, and computer vision requirements (Tab. \ref{tab:comparison} and Fig. \ref{fig:topfigure}). Camera traps capture medium to large-sized objects (128-256 pixels) \cite{he_longtailed_2023} and excel at species identification but are stationary. High-altitude drones capture very small objects (0-16 pixels) \cite{wang_surveying_2019} and are optimized for population counts across large areas but lack detail for individual ID \cite{rolland_droneswarms_2024} or behavioral analysis \cite{kholiavchenko2024kabr}.
Low-altitude drone studies capture medium-sized objects (100 pixels) \cite{kholiavchenko2024kabr}, offering a balance that enables behavioral analysis, individual ID, and tracking studies. The optimal resolution for these ecological studies is approximately 100-300 pixels per animal \cite{kline2024integrating}, which low-altitude drones provide consistently. Unlike camera traps and high-altitude drones, which typically process data in post-hoc batches, low-altitude drone applications often require real-time processing to support autonomous flight decisions \cite{kline2025wildwing, kline2024integrating}.
These fundamental differences in monitoring approaches have driven the development of specialized computer vision models for camera traps (e.g., MegaDetector \cite{beery2019efficient}) and high-altitude imagery (e.g., HerdNet \cite{delplanque2023crowd}, POLO \cite{may2024polo}). However, no pre-trained models specifically address the unique challenges of low-altitude drone imagery.

\subsection{Existing Wildlife Drone Datasets}
\begin{table*}[h]
  \centering
  \small
  \begin{tabular}{llcrc}
    \toprule
    \textbf{Dataset} & \textbf{Categories} & \textbf{Image Count} & \textbf{Size\textsuperscript{1}} & \textbf{Multi-Location\textsuperscript{2}} \\
    \midrule
    Population estimates \cite{eikelboom2019improving} & elephant, zebra, giraffe & 561 RGB & 50 & \textcolor{red}{\ding{53}} \\
    Bird detection \cite{weinstein2021global}  & bird & 24K RGB & 35 & \textcolor{green}{\checkmark} \\
    Seabird colonies \cite{Hayes_dronesbirds_2021} & albatross, penguin & 4K RGB & 300 & \textcolor{red}{\ding{53}} \\
    Conservation Drones \cite{bondi2020birdsai} & human, elephant, lion, dog & 48 thermal videos & 35 & \textcolor{green}{\checkmark} \\
    Group behavior \cite{Koger_quantifying_2024} & zebra, gazelle, waterbuck, buffalo, gelada, human & 2K RGB & 56 & \textcolor{green}{\checkmark}  \\
    KABR \cite{kholiavchenko2024kabr} & giraffe, zebra  & 130K videos & 100 & \textcolor{red}{\ding{53}}  \\
    WAID  \cite{Mou_waid_2023} & sheep, cattle, seal, camel, kiang, zebra & 14K RGB & 166 & \textcolor{red}{\ding{53}} \\
     WildLive \cite{dat2025wildlive} & zebra, giraffe, elephant & 19K RGB & 100 & \textcolor{red}{\ding{53}} \\
   WildlifeMapper  \cite{kumar2024wildlifemapper} & zebra, hartebeest, cattle, shoats, zebra & 11K RGB & --- & \textcolor{red}{\ding{53}} \\
    BuckTales \cite{naik2024bucktales} & antelope & 22.5K RGB & 1000 & \textcolor{red}{\ding{53}} \\
    \midrule
    \textbf{MMLA} (ours) & \textbf{zebra, onager, giraffe, African wild dog} & \textbf{155K RGB} & \textbf{100} & \textcolor{green}{\checkmark} \\
    \bottomrule
  \end{tabular}
  \vspace{1mm} \\
  \footnotesize\textsuperscript{1}Size indicates the typical animal size in pixels.
  \footnotesize\textsuperscript{2}Multi-Location indicates datasets capturing the same species across multiple geographic locations.
  \caption{Summary of multi-species aerial datasets. Note that MMLA has large animals (100px) and multiple locations, unlike prior work.}
  \label{tab:litreview}
\end{table*}
A critical gap exists in wildlife monitoring datasets: no multi-species, multi-location aerial footage dataset provides sufficient resolution with the same species captured across multiple environments (Tab. \ref{relatedwork}). While the global bird detection dataset \cite{weinstein2021global} contains multi-location imagery across 13 ecosystems, the animals are very small (35 pixels) and limited to avian species. Existing mammal datasets \cite{eikelboom2019improving, kholiavchenko2024kabr, Mou_waid_2023, Koger_quantifying_2024, dat2025wildlive, kumar2024wildlifemapper} contain thousands of annotated samples but are geographically limited to single locations for each species, creating detection models that may not transfer well across environments \cite{chen2018domain, vidit2023clip}.
Specialized datasets like BuckTales \cite{naik2024bucktales} and WildLive \cite{dat2025wildlive} focus on single locations or specific computer vision tasks. BuckTales provides 22.5K RGB frames of antelopes for multi-object tracking and re-identification. The WildLive dataset includes multiple species (zebras, giraffes, and elephants) but all were captured at a single location (Ol Pejeta Conservancy) \cite{dat2025wildlive}. These datasets highlight the importance of high-quality, ecologically relevant datasets tailored to specific computer vision problems but do not address the critical need for cross-environment generalization.

\subsection{Animal Detection Models}
Existing wildlife detection approaches are tailored for camera traps and high-altitude imagery, making them ill-suited for the demands of low-altitude drone studies. Camera trap pipelines like MegaDetector \cite{beery2019efficient} focus primarily on reducing false negatives to eliminate blank images \cite{orinoquia} before specialized species classification models process them. Models for high-altitude imagery use point-based approaches optimized for accurate counting from minimal pixel information \cite{may2024polo, delplanque2023crowd}.
These methods fail to meet the high precision and accurate localization requirements of automated low-altitude drone studies. Low-altitude drones often need real-time processing with high precision metrics to guide autonomous flight decisions \cite{kline2025wildwing}, where false positives can significantly impair performance by causing drones to track non-animal objects.
Our MMLA dataset aims to bridge this gap by enabling the development of robust models that can generalize across environments and species with autonomous drones.

\section{MMLA Dataset}
\label{dataset}
The MMLA dataset contains annotated drone collected for wildlife monitoring and conservation research. 
The data was collected at three locations: Ol Pejeta Conservancy and Mpala Research Centre, both in Laikipia, Kenya, and The Wilds Conservation Center, in Ohio, USA. 
The dataset contains six species: Plains zebra (\textit{Equus quagga}), Grevy's zebra (\textit{Equus grevyi}), Persian onager (\textit{Equus hemionus onager}), Reticulated giraffe (\textit{Giraffa reticulata}), Masai giraffe (\textit{Giraffa camelopardalis tippelskirchi}), and African wild dog (\textit{Lycaon pictus}).
The variety of species and environments in the MMLA dataset enables the evaluation of model generalizability in detecting animals across different settings.
The dataset includes 155,074 video frames, captured at 30 frames per second, and contains a variety of habitats (details in App. \ref{supp:data} Tab. \ref{tab:data_summary}).
Each video frame has COCO format annotations (bounding boxes with normalized coordinates).
The dataset contains 811K total annotations, which were completed manually by a team of experts using CVAT \cite{CVAT} and \texttt{kabr-tools} \cite{kline2025kabrtools}. 

The \textbf{Ol Pejeta dataset} \footnote{\href{https://huggingface.co/datasets/imageomics/mmla_opc}{https://huggingface.co/datasets/imageomics/mmla\_opc}} contains 29,268 frames of Plains zebras at Ol Pejeta Conservancy in Kenya (Jan-Feb 2025) \cite{mmla_opc}. The videos were collected using the semi-autonomous WildWing system \cite{kline2025wildwing} across seven video sessions during the 2025 WildDrone Hackathon \cite{wilddrone_eu}. 
The \textbf{Mpala dataset}\footnote{\href{https://huggingface.co/datasets/imageomics/mmla_mpala}{https://huggingface.co/datasets/imageomics/mmla\_mpala}} contains 104,062 frames of reticulated giraffes, Grevy's and Plains zebras at  Mpala Research Center in Kenya (Jan 2023) \cite{mmla_mpala}. The Mpala data was collected manually using a DJI Air 2S drone across five sessions \cite{kabr-mini-scene-videos, KABR_Raw_Videos}. 
The \textbf{Wilds dataset}\footnote{\href{https://huggingface.co/datasets/imageomics/mmla_wilds}{https://huggingface.co/datasets/imageomics/mmla\_wilds}} contains 21,744 frames of African wild dogs, Persian onagers, reticulated and Masai giraffe, and Grevy's zebras \cite{mmla_wilds, wildwingdata}. This data was collected at The Wilds in Ohio (Apr-Jul), using both DJI Mini and Parrot Anafi drones, with a combination of manually piloted flights and the semi-autonomous WildWing system \cite{kline2025wildwing, wildwingsoftware}.

\begin{table*}[t!]
\centering
\begin{tabular*}{\textwidth}{@{\extracolsep{\fill}}lrrrrr}
\toprule
Model & IoU (\%) & Precision (\%) & Recall (\%) & F1 (\%) & mAP50 (\%) \\
\midrule
YOLOv5m           & 20.88 & 34.04 & 24.94 & 27.09 & 28.79 \\
YOLOv8m           & 20.23 & 32.20 & 24.78 &26.61 & 28.00 \\
YOLOv11m           & 21.77 & 35.34 & 25.83 & 28.14 & 29.85 \\
\midrule
\textbf{YOLOv11m Finetuned on MMLA \cite{mmla_finetuned_yolo11m}} & \textbf{74.39} & \textbf{83.43} & \textbf{81.39} & \textbf{81.76} & \textbf{82.40} \\
\bottomrule
\end{tabular*}
\caption{Performance of YOLO variants on the MMLA dataset. Fine-tuning YOLOv11m yields the highest performance across all metrics.}
\label{tab:yolo_performance}
\end{table*}

\section{Evaluation}
\label{modelevaluation}

We evaluate the performance of three popular YOLO model (YOLOv5m, YOLOv8m, and YOLOv11m) for animal detection in the MMLA dataset.
The YOLO (You Only Look Once) \cite{jocher2023ultralytics} family of models represents a popular choice for animal ecology field studies due to their speed, ease of use, and accuracy \cite{kline2024characterizing}. 
We selected medium-sized variants to balance detection performance with computational efficiency, as field studies often utilize edge devices, such as laptops and Raspberry Pis, with limited memory and compute\cite{zualkernan_2022_iot, *toward_automated_monitoring_2022}.
YOLO models perform detection in a single forward pass, making them suitable for real-time applications on resource-constrained devices, such as those used for autonomous drone deployments \cite{kline2025wildwing, dat2025wildlive}.

\subsection{Pretrained YOLO}
Our initial evaluations were conducted using the pre-trained models without additional fine-tuning to assess their out-of-the-box performance on our dataset.
We extracted approximately 500 frames from each of the ten sampling sessions to ensure even coverage across the MMLA dataset for all the locations and species. 
Note: we combined the two sessions collected at the Ol Pejeta Conservancy into one session for the testing, as all videos captured Plains zebras in the same park area on consecutive days.
The evaluation subset consists of 5,073 frames, of which 1,050 from Ol Pejeta Conservancy, 1,923 from Mpala Research Center, and 2,100 from The Wilds Conservation Center.
See App. \ref{app:baseline} for detailed assessment of the baseline models' performance for each location and selected species present at most locations (zebra and giraffe).


\subsection{Fine-tuned YOLOv11m}

YOLOv11m achieved the best performance on the MMLA dataset, with an 18.6\% F1 score and 32.8\% mAP50 score, so this model was selected for fine-tuning on the dataset.
We implemented a stratified data partitioning strategy to ensure robust model generalization across our multi-location wildlife imagery dataset. We divided the data into training (58\%) and test (42\%) splits, carefully maintaining proportional representation across all three ecological zones and five target species (reticulated and Masai giraffe, Plains zebra, Grevy's zebra, Persian onager, and African wild dog.
We assigned entire video sequences to a single partition to prevent data leakage, ensuring that temporally adjacent frames never appeared in different splits when possible. Furthermore, we regularly sampled non-consecutive frames (every 10) to reduce temporal redundancy while preserving diversity in environmental conditions, animal behaviors, and camera perspectives. This approach maintained statistical consistency across partitions while ensuring independence between the training and evaluation data, with each species and location represented in at least two partitions to facilitate a thorough assessment of the model's cross-habitat performance. Please refer to our GitHub repository \footnote{\href{https://github.com/Imageomics/mmla}https://github.com/Imageomics/mmla} for instructions on generating the train/test split.
This division of training/test data by session and location ensures that the model is evaluated on unseen environments, individuals, and contexts, making it a robust benchmark for testing generalization across ecological and geographic domains. 
 
 We trained the model for 50 epochs (approximately 2 hours) on the Ohio Supercomputer Cluster, using two Tesla V100-PCIE-16GB GPUs.
 The hyperparameters and optimizers were left as the default for YOLO fine-tuning using scripts provided in the \texttt{ultralytics} library \cite{jocher2023ultralytics}. We provide scripts to reproduce our results in our GitHub repository and provide the fine-tuned weights \footnote{https://huggingface.co/imageomics/mmla} \cite{mmla_finetuned_yolo11m}. Additional training details may be found in App. \ref{app:train}.

\section{Results}
We report each model's IoU, precision, recall, F-1, and mAP50 scores in Tab.~\ref{tab:yolo_performance}.
Comparing the baseline YOLOv11m performance against the fine-tuned model reveals dramatic improvements across all metrics. 
Fine-tuning significantly improved model performance across locations and species. 
At Mpala, F1 increased by 34.3\% and mAP50 by 33.8\%, while The Wilds showed the most dramatic gains with F1 and mAP50 rising by 90.8\% and 83.5\%, respectively, indicating strong generalization to this previously underperforming site. 
These location-specific variations highlight distributional differences between fine-tuning data and habitat features.
Species-level improvements were substantial, particularly for African wild dogs (+94.9\% F1), onagers (+70.2\%), and giraffes (+30.4\%). Even zebras, which had the highest baseline performance, improved by 22.0\% F1. 

\begin{figure}
    \centering
    \includegraphics[width=1\linewidth]{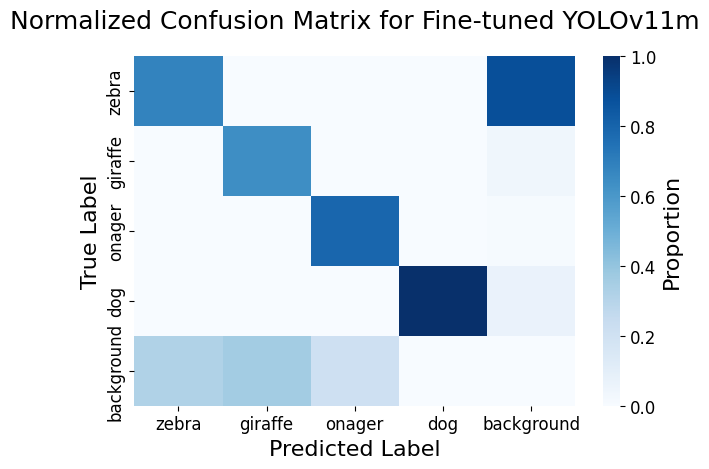}
    \caption{Prediction accuracy for fine-tuned YOLOv11m across the four classes (zebra, giraffe, onager, and dog) and background.  Results suggest that the model could be improved to more accurately detect the background and might be overfitted for African wild dog detection.}
    \label{fig:confusionmatrix}
\end{figure}

The normalized confusion matrix (Fig. \ref{fig:confusionmatrix}) for the fine-tuned YOLOv11m model reveals strong species discrimination capabilities with minimal cross-species confusion. 
 Zebras show minimal confusion with other classes due to their distinctive stripe patterns. Giraffes occasionally register false positives with background elements, likely due to their complex body patterns and tall stature blending with vegetation. Onagers demonstrate strong classification accuracy, while African wild dogs exhibit near-perfect classification due to their distinctive coat patterns and body shapes. Background elements occasionally trigger false positives across species, with slightly higher rates for giraffes and onagers, indicating successful learning of discriminative features while maintaining robustness for autonomous wildlife tracking applications.

\section{Limitations}
Our work has three main limitations that may be addressed in future work: the dataset itself, model architecture, and computational efficiency. One, our dataset has species-specific location limitations, particularly for African wild dogs, Masai giraffes, and Persian onagers, which were only present at The Wilds. Future work could combine others' datasets (Tab. \ref{tab:litreview}) with MMLA to include these species across multiple environments. Further, we combine the two zebra species and giraffe species into one class. Two, we used a baseline YOLO model without any architectural improvements specific to animal detection from drone imagery. Other works on small-object animal detection and models tailored to drone imagery (WILD-YOLO \cite{mou2024novel}, WildARe-YOLO \cite{bakana2024wildare}, and Drone-YOLO \cite{zhang2023drone}), could further improve performance on the MMLA dataset. Three, our current approach does not address deployment on resource-constrained drone platforms. Future work could incorporate energy-aware algorithms that explicitly consider edge constraints, such as dynamic pruning for edge inference \cite{oquinn2025environment}.

\section{Conclusion}
Low-altitude drone studies offer a valuable middle ground between the detailed imagery of camera traps and the broad coverage of high-altitude surveys, enabling new behavioral and movement studies. Autonomous drones combined with robust detection models can transform wildlife monitoring from a static, pre-planned process to an adaptive system that responds to animal movements in real-time \cite{kline2025wildwing}. Deploying such systems involves important tradeoffs for ecologists. Lower flight altitudes provide better resolution but may increase animal disturbance, potentially altering natural behaviors \cite{afridi2025impact, Schad_opportunities_2023}. The balance between human and automated analysis also presents challenges \cite{kline2024characterizing}. Manual analysis by trained biologists remains the reliability standard but limits study scale, while automated detection can dramatically increase processing speed but requires domain-specific training to achieve necessary reliability. The dramatic improvements achieved through fine-tuning suggest that with appropriate dataset development, AI-assisted wildlife monitoring can achieve the reliability needed for scientific applications while significantly increasing wildlife studies' temporal and spatial coverage. By contributing both the MMLA dataset and fine-tuned model weights to the research community, we hope to accelerate the development of robust computer vision solutions for wildlife conservation and establish a foundation for the next generation of intelligent, autonomous wildlife monitoring systems.

\section{Acknowledgments}
This work was supported by WildDroneEU (EU Horizon Europe MSCA grant 101071224), the U.S. National Science Foundation through the Imageomics Institute (NSF HDR Award 2118240), and the AI institute for Intelligent Cyberinfrastructure with Computational Learning in the Environment (ICICLE) (NSF grant OAC-2112606).

We thank Samuel Mutisya and William Njoroge at Ol Pejeta Conservancy for their support in facilitating the WildDrone 2025 Hackathon. We thank Jackson Miliko at the Mpala Research Centre for his support in facilitating the Kenyan Animal Behavior Recognition (KABR) project, whose data was used in this work. We thank Dan Beetem, The Wilds, and the Columbus Zoo \& Aquarium for their support in facilitating data collection. All drone missions at The Wilds was conducted under the supervision of the Director of Animal Management, with permission from The Wilds Animal Care and Use Committee.


{\small
\bibliographystyle{ieee_fullname}
\bibliography{bib}
}

\newpage
\onecolumn

\section{MMLA Dataset Details}
\label{supp:data}
\begin{table*}[h]
\centering
\small
\begin{tabular}{llcrl}
\toprule
\textbf{Location} & \textbf{Session} & \textbf{Date} & \textbf{Total Frames} & \textbf{Species} \\
\midrule
\multirow{2}{*}{Ol Pejeta Conservancy, Laikipia, Kenya} 
    & 1 & 1/31/25 & 16,726 & Plains zebra \\
    & 2 & 2/1/25 & 12,542 & Plains zebra \\
\cmidrule(lr){2-5}
    & \multicolumn{2}{r}{\textit{Subtotal:}} & \textit{29,268} & \\
\midrule
\multirow{5}{*}{Mpala Research Center, Laikipia, Kenya} 
    & 1 & 1/12/23 & 16,891 & Reticulated giraffe \\
    & 2 & 1/17/23 & 11,165 & Plains zebra \\
    & 3 & 1/18/23 & 17,940 & Grevy's zebra \\
    & 4 & 1/20/23 & 33,960 & Grevy's zebra \\
    & 5 & 1/21/23 & 24,106 & Giraffe, Plains and Grevy's zebras \\
\cmidrule(lr){2-5}
    & \multicolumn{2}{r}{\textit{Subtotal:}} & \textit{104,062} & \\
\midrule
\multirow{4}{*}{The Wilds Conservation Center, Ohio, USA} 
    & 1 & 6/14/24 & 13,749 & African wild dog \\
    & 2 & 7/31/24 & 3,436 & Masai and reticulated giraffe \\
    & 3 & 4/18/24 & 4,053 & Persian onager \\
    & 4 & 7/31/24 & 506 & Grevy's zebra \\
\cmidrule(lr){2-5}
    & \multicolumn{2}{r}{\textit{Subtotal:}} & \textit{21,744} & \\
\midrule
\multicolumn{3}{r}{\textbf{Total:}} & \textbf{155,074} & \\
\bottomrule
\end{tabular}
\caption{Frame-level summary of the MMLA dataset by location, session, date, and species.}
\label{tab:data_summary}
\end{table*}


\section{Baseline YOLO Evaluation}
\label{app:baseline}

Our evaluation of baseline YOLO models (v5m, v8m, and v11m) reveals significant performance challenges when applied to low-altitude drone imagery across different environments (Table \ref{tab:combined_f1} and Table \ref{tab:precision}). Despite achieving 49-51\% mAP50 on the standard MSCOCO dataset, these models show substantial degradation when applied to our MMLA dataset, with overall F1 scores dropping dramatically to just 16-19
Performance varied significantly across locations and species. For instance, at the Mpala location, models achieved relatively strong mAP50 scores (64-66\%) but struggled considerably at The Wilds (19-25\%) and Ol Pejeta Conservancy (26-30\%). This location-based performance disparity highlights the domain shift problem in cross-environment generalization.
Species-specific performance also showed notable patterns. Giraffe detection generally outperformed zebra detection across all models and locations, with precision scores for giraffes reaching 76-80

The newest model, YOLOv11m, achieved the best overall F1 score (18.6\%) and mAP50 (32.8\%), demonstrating incremental improvements over earlier versions. However, even this best-performing model exhibits a nearly 20-point drop in mAP50 from MSCOCO to our wildlife dataset, underscoring the significant gap between standard object detection benchmarks and the challenges of real-world low-altitude wildlife detection. This performance deficit clearly demonstrates the need for specialized models and cross-environment training approaches for autonomous wildlife drone applications.

\begin{table*}[h]
\small
\centering
\begin{tabular}{ccccccccccccc}
\toprule
\multirow{2.3}{*}{\textbf{Model}} & \multicolumn{4}{c}{\textbf{Overall F1}} & \multicolumn{4}{c}{\textbf{Giraffe F1}} & \multicolumn{4}{c}{\textbf{Zebra F1}} \\
\cmidrule(lr){2-5} \cmidrule(lr){6-9} \cmidrule(lr){10-13}
 & \textbf{All} & \textbf{Mpala} & \textbf{OPC} & \textbf{Wilds} & \textbf{All} & \textbf{Mpala} & \textbf{OPC} & \textbf{Wilds} & \textbf{All} & \textbf{Mpala} & \textbf{OPC} & \textbf{Wilds} \\
\midrule
YOLOv5m & 16.31 & \textbf{22.60} & \textbf{16.34} & 12.02 & 49.33 & 52.62 & -- & 39.98 & 10.73 & 7.10 & \textbf{18.42} & 4.03 \\
YOLOv8m & 17.01 & 20.20 & 16.08 & 15.33 & 46.05 & 49.29 & -- & 38.70 & 10.16 & 5.51 & 17.97 & 5.08 \\
YOLO11m & \textbf{18.55} & 21.86 & 15.52 & \textbf{17.59} & \textbf{50.89} & \textbf{54.59} & -- & \textbf{41.88} & \textbf{16.08} & \textbf{4.60} & 16.36 & \textbf{42.55} \\
\bottomrule
\end{tabular}
\caption{Base YOLO models F1 Scores (\%) across species and locations in MMLA.}
\label{tab:combined_f1}
\end{table*}

\begin{table*}[!htbp]
\small
\centering
\setlength{\tabcolsep}{4pt}
\begin{tabular}{cccccccccccccc}
\toprule
\multirow{2.3}{*}{\textbf{Model}} & \multirow{2.3}{*}{\makecell{\textbf{MSCOCO} \\ (mAP50)}} & \multicolumn{4}{c}{\textbf{Overall mAP50}} & \multicolumn{4}{c}{\textbf{Giraffe Precision}} & \multicolumn{4}{c}{\textbf{Zebra Precision}} \\
\cmidrule(lr){3-6} \cmidrule(lr){7-10} \cmidrule(lr){11-14}
 & & \textbf{All} & \textbf{Mpala} & \textbf{OPC}  & \textbf{Wilds} & \textbf{All} & \textbf{Mpala} & \textbf{OPC} & \textbf{Wilds} & \textbf{All} & \textbf{Mpala} & \textbf{OPC} & \textbf{Wilds} \\
\midrule
YOLOv5m & 49.0 & 31.58 & 64.88 & \textbf{29.78} & 22.17 & \textbf{76.05} & 77.91 & -- & \textbf{70.64} & 31.17 & 35.44 & \textbf{30.37} & 22.40 \\
YOLOv8m & 50.2 & 31.77 & 63.88 & 26.33 & \textbf{24.99} & 66.85 & 77.90 & -- & 47.47 & 29.67 & \textbf{36.74} & 27.05 & 37.30 \\
YOLOv11m & \textbf{51.2} & \textbf{32.83} & \textbf{65.71} & 27.06 & 19.51 & 71.53 & \textbf{80.28} & -- & 53.30 & \textbf{34.18} & 28.84 & 21.59 & \textbf{68.24} \\
\bottomrule
\end{tabular}
\caption{Precision Metrics (\%) across MMLA Species and Locations. We report the mAP50 achieved by each YOLO variant on the MSCOCO dataset \cite{coco} as a reference baseline for object detection performance \cite{jocher2023ultralytics}.}
\label{tab:precision}
\end{table*}

\section{YOLO Finetune Training Details}
\label{app:train}
\subsection{Data Splitting Strategy for Model Fine-tuning and Evaluation}
\label{sec:Suppdatasplit}

Please refer to our GitHub repository \footnote{\href{https://github.com/Imageomics/mmla}https://github.com/Imageomics/mmla} for instructions on generating the train/test split.

\begin{table}[ht]
\centering
\label{tab:dataset-split}
\begin{tabular}{p{2cm}p{7cm}p{7cm}}
\hline
\textbf{Location} & \textbf{Training Set} & \textbf{Test Set} \\
\hline
Mpala & Multiple full sessions for reticulated giraffes, Plains zebras, and Grevy's zebras, including mixed-species scenes & Separate zebra and mixed-species sessions (includes reticulated giraffes, Plains and Grevy's zebras) not used during training \\
\hline
Ol Pejeta & Full sessions of Plains zebras & Separate zebra and mixed-species sessions not used during training \\
\hline
The Wilds & 70\% of sessions for African wild dogs, giraffes (Masai and reticulated), and Persian onagers & Remaining 30\% of sessions, including additional Grevy's zebra sessions used exclusively for testing \\
\hline
\end{tabular}
\caption{MMLA Dataset Train/Test Split by Location and Species}
\end{table}

\subsection{Training and Optimization Details}
\label{supp:training}

Training was conducted for 50 epochs on two NVIDIA Tesla V100 GPUs (16~GB each) using Ultralytics YOLO v8.3.117, Python~3.10.16, and PyTorch~2.6.0 with CUDA~12.4. The total training time was 2 hour, 11 minutes. We used the Ohio Supercomputer to run our experiments.
We used the YOLO framework’s \texttt{auto} optimizer setting, which defaults to \textbf{stochastic gradient descent (SGD)} with momentum. The initial learning rate was set to 0.01 and decayed using a cosine schedule with a final learning rate fraction (\texttt{lrf}) of 0.01. Momentum was set to 0.937 and weight decay to 0.0005. A warmup phase of 3 epochs was applied, with a warmup momentum of 0.8 and a warmup bias learning rate of 0.0.

Loss weighting terms were: box loss = 7.5, class loss = 0.5, and Distribution Focal Loss (DFL) = 1.5. Mixed precision training was enabled via automatic mixed precision (AMP), and deterministic behavior was enforced using \texttt{seed = 0} and \texttt{deterministic = True}.

We applied standard YOLO augmentations, including mosaic augmentation (probability = 1.0), random horizontal flipping (0.5), HSV color jittering (hue = 0.015, saturation = 0.7, value = 0.4), random translation (0.1), and scaling (0.5). We used \textbf{RandAugment} for policy-based image augmentation and random erasing with a probability of 0.4. MixUp and CopyPaste were disabled.

Training resumed from a pretrained checkpoint and saved model weights and training plots for all runs.

\newpage

\section{YOLO11m Finetuned Results}
\label{sec:finetune_results}

\subsection{Bootstrapping}
\begin{figure}
    \centering
    \includegraphics[width=0.6\linewidth]{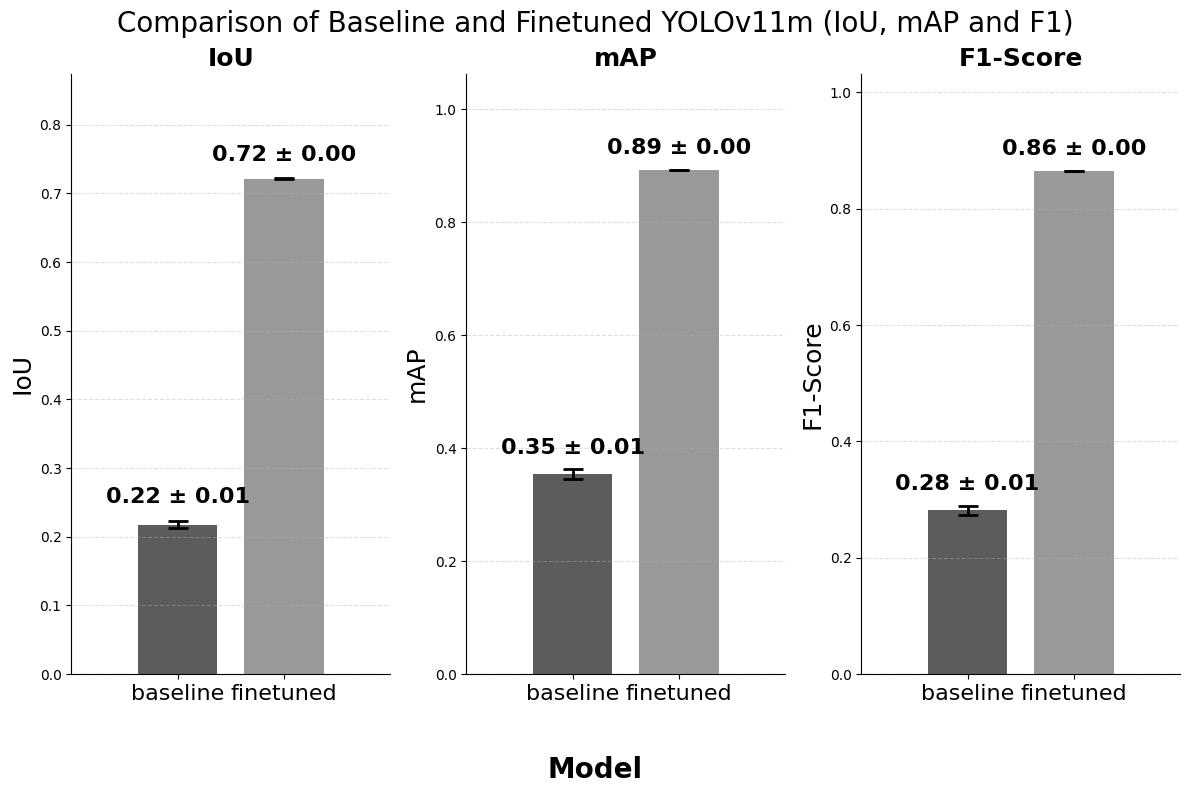}
    \caption{Bootstrapping performance of baseline and fine-tuned YOLOv11m models with confidence intervals for IoU, mAP, and F1. The finetuned model demonstrated statistically significant performance improvement across all metrics.}
    \label{fig:bootstrap}
\end{figure}

To ensure statistical significance, we performed bootstrapping with replacement on the test dataset for 1000 iterations on both the baseline and fine-tuned YOLOv11m model. We report the results with confidence intervals in Figure \ref{fig:bootstrap}. The fine-tuned model shows statically significant performance gains across IoU (23\% vs 64\%) mAP (37\% vs 89\%), and F1 (29\% vs 84\%).

\newpage
\section{YOLOv11m Baseline vs Fine-tuned Model Comparison Performance by Location and Species}
\label{supp:comp}

\begin{table}[h]
\centering
\small
\caption{Baseline Performance by Location}
\begin{tabular}{lrrrr}
\toprule
\textbf{Location} & \textbf{Precision (\%)} & \textbf{Recall (\%)} & \textbf{F1 (\%)} & \textbf{mAP50 (\%)} \\
\midrule
Mpala & 50.5 & 47.4 & 48.2 & 31.2\\
Ol Pejeta & 75.0 & 49.0 & 58.4 & 51.7\\
The Wilds & 8.7 & 3.5 & 4.5 & 5.1 \\
\midrule
Overall & 44.8 & 33.3 & 37.1 & 29.3 \\
\bottomrule
\end{tabular}
\label{tab:baseloc}
\end{table}

\begin{table}[h]
\centering
\small
\caption{Fine-Tuned Performance by Location}
\begin{tabular}{lrrrr}
\toprule
\textbf{Location} & \textbf{Precision (\%)} & \textbf{Recall (\%)} & \textbf{F1 (\%)} & \textbf{mAP50 (\%)} \\
\midrule
Mpala & 83.0 & 82.8 & 82.6 & 64.9  \\
Ol Pejeta & 55.9 & 44.4 & 49.2 & 65.8  \\
The Wilds & 95.0 & 96.6 & 95.4 & 88.6  \\
\midrule
Overall & 77.9 & 74.6 & 75.7 & 73.1  \\
\bottomrule
\end{tabular}
\label{tab:fineloc}
\end{table}

\begin{table}[h]
\centering
\small
\caption{Baseline Performance by Species}
\begin{tabular}{lrrrrr}
\toprule
\textbf{Species} & \textbf{Precision (\%)} & \textbf{Recall (\%)} & \textbf{F1 (\%)} & \textbf{mAP50 (\%)} \\
\midrule
Dog & 6.4 & 2.1 & 3.0 & 3.8  \\
Giraffe & 39.4 & 35.6 & 36.3 & 23.2\\
Onager & 26.3 & 11.7 & 14.1 & 15.9\\
Zebra & 57.0 & 47.7 & 50.7 & 36.5 \\
\midrule
Overall & 32.3 & 24.3 & 26.0 & 19.9 \\
\bottomrule
\end{tabular}
\label{tab:basespecies}
\end{table}

\begin{table}[h]
\centering
\small
\caption{Fine-Tuned Performance by Species}
\begin{tabular}{lrrrr}
\toprule
\textbf{Species} & \textbf{Precision (\%)} & \textbf{Recall (\%)} & \textbf{F1 (\%)} & \textbf{mAP50 (\%)} \\
\midrule
Dog & 96.6 & 99.6 & 97.9 & 90.8\\
Giraffe & 100.0 & 50.0 & 66.7 & 71.2 \\
Onager & 92.9 & 81.5 & 84.3 & 79.2 \\
Zebra & 74.8 & 71.7 & 72.8 & 64.1\\
\midrule
Overall & 91.1 & 75.7 & 80.4 & 76.3 \\
\bottomrule
\end{tabular}
\label{tab:finespecies}
\end{table}

\newpage
\section{Performance Metrics for Wildlife Monitoring Approaches}


For autonomous low-altitude drone systems like WildWing \cite{kline2025wildwing}, precision metrics are crucial, more so than in traditional wildlife monitoring approaches. In these systems, false positives significantly impair performance by causing the drone to adjust its flight path or focus on non-animal objects.
When an autonomous drone relies on detection algorithms to identify and track animals in real-time, each false positive potentially redirects valuable flight time and battery capacity away from actual wildlife subjects. For example, a drone following what it incorrectly identified as an animal could lose track of the actual subjects of interest, resulting in failed data collection missions and wasted resources.
Additionally, Intersection over Union (IoU) metrics are particularly important for autonomous tracking, as precise bounding box predictions directly influence flight control decisions. A detection model with high IoU scores provides more accurate animal positioning data, allowing the drone to maintain optimal following distance and camera angles. Studies have shown that IoU values of at least 0.5 (mAP50) are necessary for stable autonomous tracking \cite{kline2024integrating}.
Autonomous drones also require faster inference times than post-hoc batch processing systems. While camera trap and high-altitude drone footage can be processed hours or days after collection, autonomous tracking requires near real-time analysis (typically <100ms per frame) to make timely flight control decisions. This creates a challenging optimization problem where precision, IoU accuracy, and inference speed must all be balanced.

The following metrics are critical for evaluating detection models for autonomous wildlife drones:

\begin{itemize}
    \item \textbf{Precision}: The proportion of positive identifications that were actually correct. High precision minimizes wasteful flight path adjustments.
    \item \textbf{mAP50}: Mean Average Precision with IoU threshold of 0.5, measuring how accurately the model localizes animals within frames.
    \item \textbf{Inference Time}: Processing time per frame, which must be sufficiently fast for real-time flight decisions.
    \item \textbf{Cross-Environment Performance}: How well precision and mAP50 metrics transfer when the model is deployed in new environments not seen during training.
\end{itemize}

Our baseline model evaluations in this study prioritize these metrics, particularly focusing on precision and cross-environment performance as the most critical for autonomous wildlife tracking applications.

\end{document}